\documentclass[10pt,twocolumn,letterpaper]{article}

\usepackage{cvpr}              
\usepackage{amsmath}
\usepackage{float}
\usepackage{multirow}
\usepackage{adjustbox} 
\usepackage{booktabs}
\usepackage{tabularx}
\usepackage{cuted}
\usepackage{graphicx}
\usepackage{subcaption}  
\usepackage{pifont}   
\usepackage{placeins}

\definecolor{cvprblue}{rgb}{0.21,0.49,0.74}
\usepackage[pagebackref,breaklinks,colorlinks,allcolors=cvprblue]{hyperref}
\usepackage{booktabs}
\usepackage[table]{xcolor}

\def\paperID{18} 
\def\confName{CVPR}
\def\confYear{2026}

\title{ProDrive: Proactive Planning for Autonomous Driving via Ego-Environment Co-Evolution}

\author{
Chuyao Fu$^{1}$\quad
Shengzhe Gan$^{1}$ \quad
Zhuoli Ouyang$^{1}$ \quad
Yuhan Rui$^{1}$ \\
Xiaowei Chi$^{2}$ \quad
Sirui Han$^{2}$ \quad
Jiankun Wang$^{1*}$ \quad
Hong Zhang$^{1*}$\\[0.5em]
$^{1}$Southern University of Science and Technology\\
$^{2}$Hong Kong University of Science and Technology\\
{\tt\small \{wangjk, hzhang\}@sustech.edu.cn}\\
{\small $^{*}$Corresponding authors}
}

\begin{document}
\maketitle

\begin{strip}
    \centering
    \includegraphics[width=\textwidth]{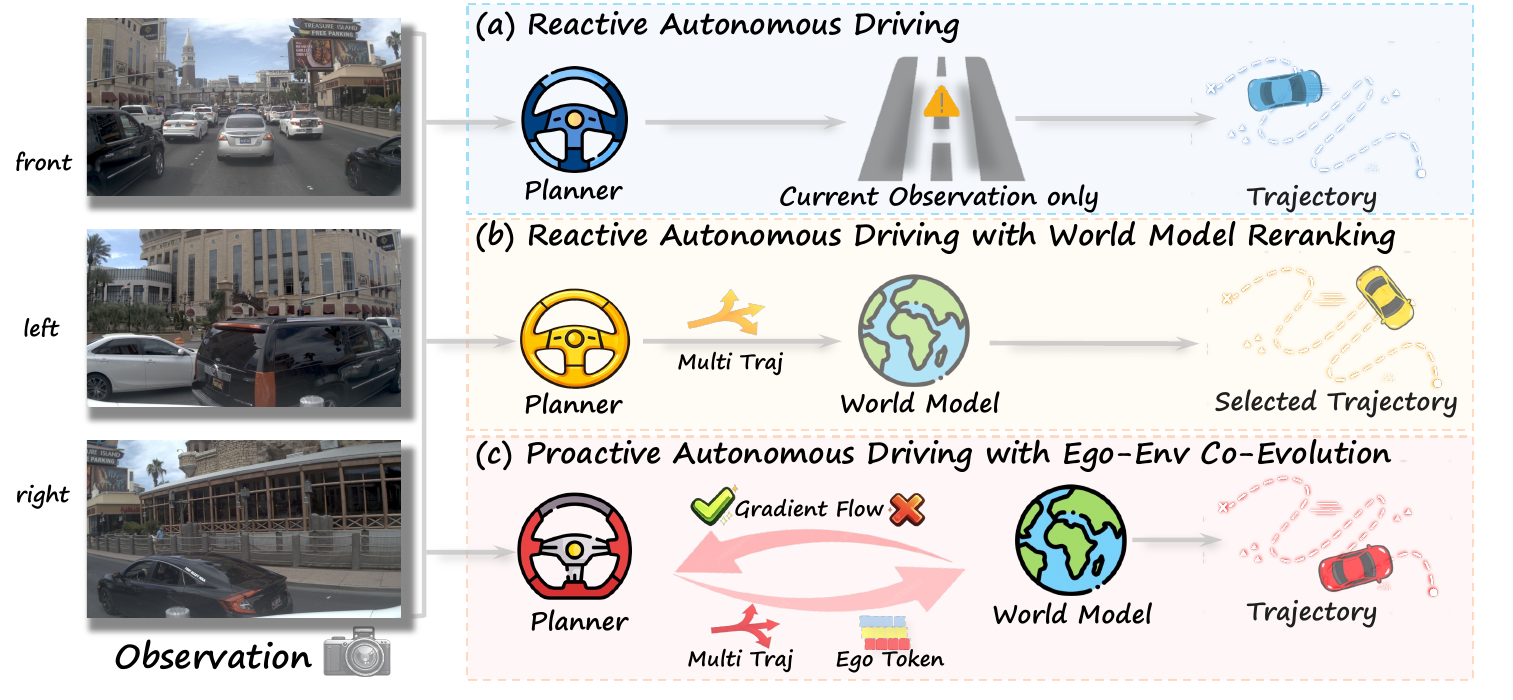}
    \captionof{figure}{\textbf{From reactive to proactive autonomous driving.}
    (a) Conventional end-to-end planners are \emph{reactive}, generating trajectories mainly from the current observation without explicitly modeling future scene evolution.
    (b) Some recent methods use a world model for trajectory reranking, but the planner and world model remain loosely coupled, limiting direct planner-side benefit from future reasoning.
    (c) In contrast, \textbf{ProDrive} enables \emph{proactive} driving through \textbf{ego-environment co-evolution}: the planner provides ego tokens for planner-aware future prediction, while the world model evaluates candidate trajectories and guides the planner through end-to-end gradient feedback.}
    \label{fig:placeholder}
\end{strip}

\begin{abstract}

End-to-end autonomous driving planners typically generate trajectories from current observations alone. However, real-world driving is highly dynamic, and such reactive planning cannot anticipate future scene evolution, often leading to myopic decisions and safety-critical failures. We propose \textbf{ProDrive}, a world-model-based proactive planning framework that enables \textbf{ego-environment co-evolution} for autonomous driving. ProDrive jointly trains a query-centric trajectory planner and a bird's-eye-view (BEV) world model end-to-end: the planner generates diverse candidate trajectories and planning-aware ego tokens, while the world model predicts future scene evolution conditioned on them. By injecting planner features into the world model and evaluating all candidates in parallel, ProDrive preserves end-to-end gradient flow and allows future outcome assessment to directly shape planning. This bidirectional coupling enables proactive planning beyond current-observation-driven decision-making. Experiments on NAVSIM v1 show that ProDrive outperforms strong baselines in both safety and planning efficiency, while ablations validate the effectiveness of the proposed ego-environment coupling design.
\end{abstract}    
\section{Introduction}
\label{sec:intro}

Safe and effective autonomous driving requires both anticipating future scene evolution and making decisions from raw sensory observations. Earlier planning-conditioned trajectory prediction works observed that the future behaviors of surrounding agents should be inferred conditioned on the ego vehicle's intended goal or candidate plan, rather than from historical observations alone, enabling interaction-aware ``what-if'' reasoning \cite{song2021learningpredictvehicletrajectories,Song_2020,rhinehart2019precogpredictionconditionedgoals}. More recently, planning-oriented end-to-end autonomous driving has emerged as a dominant paradigm, directly mapping raw sensor observations to planning outputs through jointly optimized networks \cite{UniAD,transfuser,fu2025orion,LTF,PARA_Drive,VADv2,guo2025ipad,zhao2025diffe2erethinkingendtoenddriving,sun2024sparsedriveendtoendautonomousdriving}. However, these two lines of research have not yet been fully unified: planning-conditioned prediction remains largely prediction-centric, while most end-to-end driving systems still generate trajectories primarily from current observations.

In parallel, world models have recently emerged as a promising mechanism to inject future reasoning into autonomous driving systems. Existing efforts have incorporated world models in several different forms. Some train high-fidelity pixel-level world models as data engines to generate rich synthetic experiences, especially for long-tail scenarios \cite{DriveDreamer,GAIA1,zhang2025eponaautoregressivediffusionworld,zhao2024drivedreamer2llmenhancedworldmodels,ma2024unleashinggeneralizationendtoendautonomous,huang2024subjectdrivescalinggenerativedata}. Others treat world models as environment simulators for policy learning, leveraging imagined rollouts to train driving policies with reinforcement learning \cite{li2024think2driveefficientreinforcementlearning,goff2025learning}. More closely related to planning, world models have also been adopted as external trajectory evaluation modules, allowing online assessment of candidate trajectories and improving performance through reranking or selection \cite{li2025endtoenddrivingonlinetrajectory,zheng2025world4driveendtoendautonomousdriving}.

Despite these advances, world models in most existing autonomous driving systems still play an auxiliary role rather than serving as integral components of the planner itself. In many cases, they function as data generators, simulators or post-hoc evaluators, while trajectory generation remains fundamentally driven by the current observation. Even recent future-aware driving frameworks that introduce prediction or imagination into end-to-end driving \cite{zhang2025futureawareendtoenddrivingbidirectional,zhao2025forecastingplanningpolicyworld,zeng2025futuresightdrivethinkingvisuallyspatiotemporal,li2026imagidriveunifiedimaginationandplanningframework} often lack the ability to tightly couple planner-side decision representations with predicted future scene evolution. As a result, planning remains largely \emph{reactive}: future reasoning may assist candidate evaluation, but it rarely acts as a first-class training signal that directly shapes the planner's internal decision process. This limitation is particularly critical in autonomous driving, where safe and effective behavior depends on anticipating interactions and uncertainties in an evolving multi-agent environment.

To address this gap, we propose \textbf{ProDrive}, a world-model-based proactive planning framework that jointly models ego intention and environment evolution for autonomous driving. ProDrive consists of two tightly coupled components: an \textbf{Ego Module}, which performs structured planning from multi-view images with query-centric representations, and an \textbf{Environment Module}, which predicts future bird's-eye-view (BEV) scene evolution and evaluates candidate trajectories. Unlike conventional planning-and-reranking pipelines, ProDrive couples these components directly at the feature and optimization levels. The Ego Module provides planner-refined ego tokens to guide future environment modeling, while the Environment Module evaluates candidate trajectories in parallel and propagates future-aware gradients back to the planner. Consequently, the planner is no longer optimized solely from current observations, but is explicitly shaped by predicted future scene evolution, enabling proactive planning through ego-environment co-evolution.

Experiments on NAVSIM v1 \cite{dauner2024navsim} show that ProDrive consistently outperforms strong reactive-planning baselines, yielding improvements across safety and planning-related metrics. Qualitative results further reveal that ProDrive can anticipate future scene evolution in complex scenarios and produce more foresighted plans. Ablation studies confirm the effectiveness of the proposed Ego-Environment coupling and the benefit of jointly modeling ego intention and environment evolution. Our main contributions are summarized as follows:
\begin{itemize}
    \item We propose ProDrive, a world-model-based proactive planning framework for autonomous driving that explicitly models future observations to improve planning quality and safety.
    \item We develop an Ego-Environment collaborative architecture that tightly integrates the trajectory planner and the world model, enabling deep interaction between ego intention modeling and environment evolution modeling.
    \item We demonstrate on NAVSIM v1 that ProDrive consistently improves safety and planning efficiency over strong baselines, and further validate the effectiveness of the proposed Ego-Environment collaborative design through ablation studies.
\end{itemize}

\section{Related Work}
\label{sec:related}

\subsection{Trajectory Prediction for Planning}

Trajectory prediction has long been a core research direction in autonomous driving \cite{Madjid_2026}, as predicted future trajectories of surrounding agents are typically used by downstream planners to filter unsafe maneuvers and assess candidate actions. Representative methods such as S2TNet~\cite{chen2021s2tnet} and MultiPath++~\cite{varadarajan2021multipathefficientinformationfusion} have substantially improved forecasting quality by modeling scene context, agent interactions and multimodal futures. These works establish trajectory prediction as an important interface between perception and planning, but they generally formulate the problem as forecasting surrounding-agent motion, with planning treated as a downstream consumer of the predicted results.

A closely related line of work began to narrow the gap between prediction and planning by conditioning forecasting on ego intention or candidate plans. PRECOG~\cite{rhinehart2019precogpredictionconditionedgoals}, Trajectron++~\cite{salzmann2021trajectrondynamicallyfeasibletrajectoryforecasting}, PiP~\cite{Song_2020} and PRIME~\cite{song2021learningpredictvehicletrajectories} showed that surrounding-agent futures should depend not only on history, but also on the ego vehicle's goal, candidate trajectory or planning constraints. However, these methods remain largely prediction-centric: planning information is introduced to improve conditional forecasting or candidate evaluation, while the planner itself is still not directly shaped by future-scene reasoning. In contrast, ProDrive makes future reasoning an integral part of planning itself by performing ego-conditioned future BEV rollouts and propagating proactive gradients back to the planner through ego-environment coupling.

\subsection{End-to-End Autonomous Driving}

End-to-end autonomous driving aims to learn a direct mapping from raw sensor inputs to planning outputs, bypassing the traditional modular pipeline of perception, prediction, and planning. Early works such as TransFuser~\cite{transfuser} introduced transformer-based multi-modal fusion to combine camera images and LiDAR representations at multiple resolutions, achieving strong imitation learning performance on the CARLA \cite{dosovitskiy2017carla} benchmark. Its camera-only variant, Latent TransFuser (LTF)~\cite{LTF}, demonstrated that competitive driving performance could be attained without LiDAR by substituting positional encodings for the LiDAR branch.

A pivotal advance came with UniAD~\cite{UniAD}, which proposed a planning-oriented philosophy that unifies perception, prediction, and planning within a single query-based framework. By using learnable queries as flexible interfaces connecting all sub-tasks, UniAD established that joint optimization toward the ultimate planning objective significantly reduces compounding errors inherent in sequential modular pipelines. PARA-Drive~\cite{PARA_Drive} further improved upon this paradigm by parallelizing the perception, prediction, and planning modules rather than executing them sequentially, achieving nearly 3$\times$ speedup while maintaining state-of-the-art accuracy.

VADv2~\cite{VADv2} advanced the vectorized scene representation paradigm by formulating planning as a probabilistic distribution over actions rather than deterministic trajectory regression, enabling the model to capture the inherent multi-modality of driving decisions. These methods collectively demonstrate the rapid progress in E2E driving; however, they remain predominantly \emph{reactive}, generating trajectories conditioned solely on the current observation without explicitly modeling how the environment will evolve in response to the ego vehicle's actions.

\subsection{World Models for Autonomous Driving}

World models provide a mechanism for anticipatory reasoning by predicting future environment states. In autonomous driving, existing work mainly falls into two categories: \emph{generative} world models and \emph{structured} world models. Generative approaches, such as GAIA-1~\cite{GAIA1}, DriveDreamer~\cite{DriveDreamer}, and ADriver-I~\cite{ADriver_I}, focus on synthesizing realistic future driving videos or frames from multimodal inputs. While visually compelling, these methods operate in high-dimensional pixel spaces, making them costly and difficult to tightly integrate with downstream planning.

An alternative line of work explores structured world models that predict compact, task-relevant representations. LAW~\cite{LAW}, for example, learns a latent world model that forecasts future scene features conditioned on ego actions, using self-supervision to improve feature learning and action prediction. More recent efforts have further connected future modeling with end-to-end planning. SeerDrive~\cite{zhang2025futureawareendtoenddrivingbidirectional} predicts future BEV representations to support planning, PWM~\cite{zhao2025forecastingplanningpolicyworld} unifies forecasting and planning in a policy world model, FutureSightDrive~\cite{zeng2025futuresightdrivethinkingvisuallyspatiotemporal} uses visual spatio-temporal reasoning for future-aware planning, and ImagiDrive~\cite{li2026imagidriveunifiedimaginationandplanningframework} couples a driving agent with a scene imaginer in an imagination-and-planning loop.

Our method is most closely related to the structured world modeling direction, but differs in one key aspect. Rather than conditioning the world model on generic latent features or decoupled action tokens, we directly inject the planner's dynamically refined trajectory tokens into a BEV world model. This tighter planner--world-model coupling provides richer planning-aware semantics for future prediction, enabling candidate-specific future rollouts and more informative reward-based reranking.

\begin{figure*}[htbp]
    \centering
    \includegraphics[width=\linewidth]{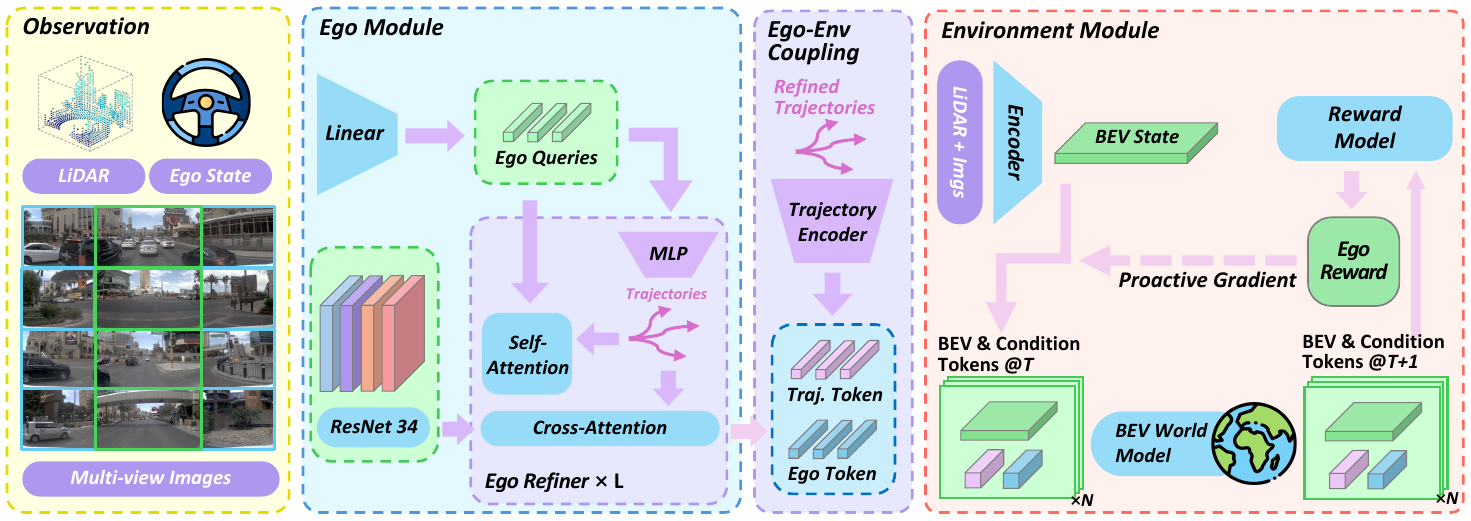}
    \caption{\textbf{Overview of ProDrive.}
    Given multi-view images, LiDAR, and ego state, the \textbf{Ego Module} refines learnable ego queries through \(L\) Ego Refiner layers to obtain ego tokens, from which candidate trajectories are decoded and transformed into trajectory tokens. Conditioned on these tokens and the current BEV state, the \textbf{Environment Module} performs recurrent BEV future prediction and reward-based trajectory evaluation. Through ego token injection and end-to-end gradient feedback, this bidirectional coupling enables ego-environment co-evolution and allows ProDrive to learn proactive planning from anticipated future scene dynamics.}
    \label{fig:placeholder}
\end{figure*}
\section{Method}
\label{sec:method}

ProDrive consists of two tightly coupled components: an \textbf{Ego Module} (\ref{sec:ego_module}), a query-centric planner that refines learnable ego tokens to generate diverse candidate trajectories, and an \textbf{Environment Module} (\ref{sec:environment_module}), a BEV world model that predicts future scene states and evaluates these candidates. We further introduce an \textbf{Ego-Environment Coupling} mechanism (\ref{sec:ego_environment_coupling}) that connects the two modules, enabling bidirectional interaction and co-evolution between ego planning and environment dynamics.

\subsection{Problem Formulation}
\label{sec:formulation}

Given multi-view camera images $\mathcal{I} = \{I_1, \ldots, I_C\}$ from $C$ cameras, LiDAR point cloud $\mathcal{P}$, and ego vehicle status $\mathbf{s}_{\text{ego}} \in \mathbb{R}^{d_s}$ (velocity, acceleration, yaw rate), the goal is to predict a future trajectory $\hat{\boldsymbol{\tau}} = \{(\hat{x}_t, \hat{y}_t, \hat{\theta}_t)\}_{t=1}^{T}$ consisting of $T$ waypoints in the ego-centric coordinate frame, where each waypoint specifies position $(x, y)$ and heading $\theta$.

\subsection{Ego Module: Query-Centric Planner}
\label{sec:ego_module}

The Ego Module is a query-centric planner built on a BEVFormer-based architecture \cite{li2022bevformerlearningbirdseyeviewrepresentation}. It encodes multi-view images with a ResNet-34 backbone \cite{he2015deepresiduallearningimage} and projects ego state into ego-conditioned features, which are added to a set of learnable ego tokens \(\mathbf{Q}^{(0)} \in \mathbb{R}^{(K \cdot T) \times d}\). The resulting planning tokens are iteratively refined to produce trajectory-aware representations for diverse candidate trajectories.

\paragraph{Iterative Cross-Attention Refinement.}
The ego tokens are refined through $L$ cascaded refinement stages with shared parameters. At each stage $l$, the process consists of two steps:

\noindent\textit{Step 1: Trajectory Decoding.} An MLP head decodes each ego token into a trajectory waypoint:
\begin{equation}
    \boldsymbol{\tau}^{(l)} = \text{MLP}_{\text{traj}}(\mathbf{Q}^{(l)}) \in \mathbb{R}^{K \times T \times 3},
\end{equation}
yielding $K$ trajectory proposals at refinement stage $l$.

\noindent\textit{Step 2: Feature Refinement via Cross-Attention.} The decoded trajectory positions $\boldsymbol{\tau}^{(l)}$ serve as reference points for deformable cross-attention back into the multi-view image features. The ego tokens are updated by attending to image features at spatial locations corresponding to each trajectory waypoint:
\begin{equation}
    \mathbf{Q}^{(l+1)} = \text{EgoRefiner}(\mathbf{Q}^{(l)}, \boldsymbol{\tau}^{(l)}, \mathbf{F}_{\text{img}}),
\end{equation}
where $\text{EgoRefiner}$ consists of deformable self-attention among the ego tokens, followed by deformable cross-attention with multi-view image features using the trajectory waypoints as 3D reference points projected onto each camera view.

This iterative refinement ensures that the ego tokens progressively encode richer spatial context from the scene at locations relevant to each candidate trajectory. After $L$ stages, the final trajectory candidates $\{\boldsymbol{\tau}^k\}_{k=1}^K = \boldsymbol{\tau}^{(L)}$ and the refined ego tokens $\mathbf{Q}^{(L)} \in \mathbb{R}^{(K \cdot T) \times d}$ are produced.

\paragraph{Ego-module training.}
The Ego Module is trained with trajectory, scoring, and auxiliary perception losses. For trajectory generation, we apply a winner-take-all regression loss over all refinement stages together with a diversity regularizer:
\begin{equation}
\mathcal{L}_{\text{traj}}
=
\sum_{l=1}^{L}
\gamma^{L-l}
\left(
\min_{k}\frac{1}{T}\left\|\boldsymbol{\tau}^{k,(l)}-\boldsymbol{\tau}^{*}\right\|_{1}
+
\lambda_{\text{div}}\mathcal{L}_{\text{div}}^{(l)}
\right),
\end{equation}
where \(\boldsymbol{\tau}^{k,(l)}\) is the \(k\)-th proposal at stage \(l\), \(\boldsymbol{\tau}^{*}\) is the expert trajectory, and \(\mathcal{L}_{\text{div}}^{(l)}\) encourages proposal diversity. We further supervise the planner-side scorer using online proposal-wise PDM targets and auxiliary labels, including key-agent states, validity indicators, and ego-area occupancy:
\begin{equation}
\begin{aligned}
\mathcal{L}_{\text{score}}
=
&\;
\lambda_{\text{final}}
\,\mathrm{BCE}\!\left(\hat{s}^{\text{ego}}, s^{*}\right)
+
\lambda_{\text{valid}}
\,\mathrm{BCE}\!\left(\hat{\mathbf{v}}, \mathbf{v}\right) \\
&+
\lambda_{\text{state}}
\,\left\|\hat{\mathbf{g}}-\mathbf{g}\right\|_{1}
+
\lambda_{\text{area}}
\,\mathrm{BCE}\!\left(\hat{\mathbf{a}}, \mathbf{a}\right),
\end{aligned}
\end{equation}
where \(s^{*}\) is the online PDM target and the \(L_1\) term is evaluated only on valid agent entries. In addition, we retain auxiliary BEV perception losses from the ego queries,
\begin{equation}
\mathcal{L}_{\text{aux}}
=
\lambda_{\text{bev}}
\,\mathrm{CE}\!\left(\hat{\mathbf{M}}^{\text{ego}}, \mathbf{M}^{*}\right)
+
\lambda_{\text{cls}}\mathcal{L}_{\text{det-cls}}
+
\lambda_{\text{box}}\mathcal{L}_{\text{det-box}},
\end{equation}
which stabilize the ego token representation and improve candidate generation. Together, these objectives encourage accurate and diverse proposals, planning-aware scoring, and robust ego token learning.

\subsection{Environment Module: BEV World Model}
\label{sec:environment_module}

To assess each candidate under future interaction, we employ a separate camera-LiDAR backbone to produce an initial BEV state $\mathbf{B}_0$ following TransFuser~\cite{transfuser}. For each proposal $\boldsymbol{\tau}^k$, an action token $\mathbf{a}^k$ encodes trajectory geometry and ego kinematics. 



\paragraph{Recurrent Future Prediction.}
The world model predicts future BEV states over $N$ iterations. At each iteration $i$, the input sequence is constructed as: 
\begin{equation} \mathbf{F}_i = [\mathbf{a}^k_i;\; \mathbf{s}^k_i;\; \mathbf{B}^k_i] \in \mathbb{R}^{(hw + 2) \times d}, 
\end{equation} 
where $\mathbf{a}^k_i$ is the action token, $\mathbf{s}^k_i$ is the ego token (further illustrated in \ref{sec:ego_environment_coupling}), and $\mathbf{B}^k_i$ is the current BEV feature. This sequence is augmented with learned positional embeddings and processed by a Transformer encoder, which can be formulated as an equation:

\begin{equation} 
[\mathbf{a}^k_{i+1};\; \hat{\mathbf{s}}^k_i;\; \mathbf{B}^k_{i+1}] = \text{WorldModel}(\mathbf{F}_i + \mathbf{P}_{\text{scene}}). 
\end{equation} 

The output yields the predicted future action token $\mathbf{a}^k_{i+1}$, an enriched state token $\hat{\mathbf{s}}^k_i$, and the predicted future BEV state $\mathbf{B}^k_{i+1}$.

\paragraph{Trajectory Selection.}
For each candidate, we aggregate multi-step BEV features and action tokens into a compact reward representation. Two heads predict an imitation reward supervised by the distance to the expert trajectory and simulation-oriented rewards for collision avoidance, drivable-area compliance, progress, time-to-collision, and comfort. The final score is

\begin{equation}
\begin{split}
R^k &= w_0 \log R_{\text{im}}^k + w_1 \log S_{\text{NC}}^k + w_2 \log S_{\text{DAC}}^k \\
&\quad + w_3 \log(5S_{\text{TTC}}^k + 2S_{\text{C}}^k + 5S_{\text{EP}}^k)
\end{split}
\end{equation}
where safety-related terms act as strong constraints while efficiency and comfort are softly traded off. Both during training and inference, the trajectory with the highest score is selected via \emph{argmax}.

\paragraph{Environment-module training.}
The Environment Module is trained with reward supervision and semantic BEV prediction. For trajectory-level reward learning, it predicts both imitation-style and simulation-style rewards. The imitation reward is supervised by a soft target derived from trajectory proximity to the expert,
\begin{equation}
q^{k}
=
\frac{\exp\!\left(-\|\boldsymbol{\tau}^{k}-\boldsymbol{\tau}^{*}\|_{2}\right)}
{\sum_{j}\exp\!\left(-\|\boldsymbol{\tau}^{j}-\boldsymbol{\tau}^{*}\|_{2}\right)},
\qquad
\mathcal{L}_{\text{im}}
=
-\sum_{k} q^{k}\log \hat{r}_{\text{im}}^{k},
\end{equation}
while the simulation reward is supervised using precomputed simulator metrics from the nearest anchor trajectory,
\begin{equation}
\pi(k)=\arg\min_{j}\left\|\boldsymbol{\tau}^{k}-\bar{\boldsymbol{\tau}}^{j}\right\|_{2},
\qquad
\mathcal{L}_{\text{sim}}
=
\mathrm{BCE}\!\left(
\hat{\mathbf{r}}_{\text{sim}}^{k},
\mathbf{r}_{\text{sim}}^{\pi(k)}
\right).
\end{equation}
To further couple the ego module and the environment module, we align the planner-side score with the normalized world-model score,
\begin{equation}
\mathcal{L}_{\text{align}}
=
\frac{1}{BK}
\sum_{b=1}^{B}\sum_{k=1}^{K}
\left\|
\sigma(\ell^{\text{ego}}_{b,k})-\tilde{R}^{\text{wm}}_{b,k}
\right\|_{2}^{2},
\end{equation}
and define
\begin{equation}
\mathcal{L}_{\text{reward}}
=
\lambda_{\text{im}}\mathcal{L}_{\text{im}}
+
\lambda_{\text{sim}}\mathcal{L}_{\text{sim}}
+
\lambda_{\text{align}}\mathcal{L}_{\text{align}}.
\end{equation}
In addition, to learn planning-relevant scene dynamics, we supervise both the current and future BEV semantic maps predicted by the world model:
\begin{equation}
\mathcal{L}_{\text{wm}}
=
\lambda_{\text{cur}}
\,\mathrm{Focal}\!\left(\hat{\mathbf{M}}_{0}, \mathbf{M}_{0}^{*}\right)
+
\lambda_{\text{fut}}
\,\mathrm{Focal}\!\left(\hat{\mathbf{M}}_{f}, \mathbf{M}_{f}^{*}\right),
\end{equation}
where the future target is constructed in a proposal-conditioned manner by rendering the ego box at the sampled proposal position on the future BEV semantic canvas. Together, these objectives encourage the Environment Module to produce informative reward signals and planning-aware future scene predictions.

\subsection{Ego-Environment Coupling}
\label{sec:ego_environment_coupling}

ProDrive couples the Ego and Environment modules in a bidirectional manner. The Ego Module improves future prediction through \textbf{ego token injection}, while the Environment Module guides the Ego Module through end-to-end optimization, enabling proactive planning from predicted future scene evolution.

\paragraph{Ego Token Injection.}
A key challenge in coupling a planner with a world model is that future prediction may lose the planner's internal decision semantics. Existing methods~\cite{LAW} often condition the world model on generic latent features or raw trajectory coordinates, which discard the richer context accumulated during planning. To address this issue, we inject refined planner features directly into the world model. Specifically, the planner output \(\mathbf{Q}^{(L)} \in \mathbb{R}^{(K \cdot T) \times d}\) is reshaped into \(\mathbf{Q}_{\text{plan}} \in \mathbb{R}^{K \times T \times d}\). At world model iteration \(i\), the aligned feature at timestep \(t_i\) is projected into an ego token:
\begin{equation}
    \mathbf{s}^k_i = \text{MLP}_{\text{state}}(\mathbf{Q}_{\text{plan}}[k, t_i, :]) \in \mathbb{R}^d,
\end{equation}
where \(t_i\) denotes the trajectory timestep corresponding to iteration \(i\). This allows future prediction to depend not only on \emph{where} the ego vehicle will move, but also on \emph{why} the planner proposes that motion.

\paragraph{Joint Training Objective.}
We train ProDrive end-to-end with a multi-task objective that jointly supervises proposal generation in the Ego Module, proposal evaluation in the Environment Module, and the coupling between the two:
\begin{equation}
\mathcal{L}
=
\mathcal{L}_{\text{traj}}
+
\mathcal{L}_{\text{score}}
+
\mathcal{L}_{\text{reward}}
+
\mathcal{L}_{\text{wm}}
+
\mathcal{L}_{\text{aux}},
\end{equation}
where the scalar coefficients are omitted in the main text for clarity.

\section{Experiment}
\label{sec:experiment}
Our experiments are designed to answer two key questions: \textbf{(1)} Can ProDrive generate planning behaviors that are safe, effective, and comfortable in challenging driving environments? \textbf{(2)} Do the proposed components work collaboratively as intended, and how do different design choices affect the overall system performance?
\subsection{Benchmark}

We evaluate \textbf{ProDrive} on the \textit{NAVSIM} \cite{dauner2024navsim} benchmark, which is built upon \textit{nuPlan} \cite{caesar2021nuplan} and is designed to emphasize challenging planning scenarios. Specifically, NAVSIM is constructed from driving logs that were first downsampled and condensed by OpenScene \cite{peng2023openscene3dsceneunderstanding}, and then resampled to prioritize difficult cases while reducing simple scenarios such as straight-line driving. The resulting dataset contains two splits: \textit{Navtrain}, which includes 1,192 scenarios for training and validation, and \textit{Navtest}, which contains 136 scenarios for evaluation. Compared with datasets collected in simpler and slower driving settings, NAVSIM places stronger emphasis on interactive and challenging scenarios, making it a more suitable benchmark for evaluating planning-oriented autonomous driving systems.

Following the standard NAVSIM protocol, we report the \textit{Predictive Driver Model Score} (\textbf{PDMS}) as the primary evaluation metric. Unlike earlier end-to-end driving metrics that mainly measure deviation from human expert trajectories, NAVSIM adopts a more practical evaluation scheme that better reflects real driving quality. Specifically, PDMS is computed from five factors: \textit{No At-Fault Collision} (\textbf{NC}), \textit{Drivable Area Compliance} (\textbf{DAC}), \textit{Time-to-Collision} (\textbf{TTC}), \textit{Comfort} (\textbf{C}), and \textit{Ego Progress} (\textbf{EP}). The overall score is defined as
\begin{equation}
\mathrm{PDMS} = \mathrm{NC} \times \mathrm{DAC} \times \frac{5 \cdot \mathrm{EP} + 5 \cdot \mathrm{TTC} + 2 \cdot \mathrm{C}}{12}.
\end{equation}
This metric jointly evaluates safety, rule compliance, efficiency, and ride comfort, providing a comprehensive measure of planning performance.

\subsection{Implementation Details}
\label{sec:impl}

Unless otherwise stated, all experiments follow the default training recipe in our released implementation. ProDrive is implemented in PyTorch and PyTorch Lightning.  Both modules use ResNet-34 image backbones, and the Environment Module additionally employs a ResNet-34 LiDAR backbone. The agent consumes one synchronized sensor slice indexed by \texttt{[3]} from the NAVSIM history window, including four surround-view cameras (front camera, left camera, right camera, and bottom camera) and one LiDAR sweep. We use a 4\,s planning horizon with a 0.5\,s sampling interval, resulting in $T=8$ trajectory waypoints. The query-centric planner produces $K=64$ candidate trajectories, and the full model is trained with both the world-model branch and the ego token injection enabled unless otherwise specified.

For training, we train the model from scratch and all parameters remain trainable. Optimization is performed with Adam using two parameter groups: a learning rate of $10^{-4}$ for the Ego Module and $10^{-5}$ for the Environment Module, which stabilizes joint optimization when coupling proposal generation and future prediction. Training runs for 15 epochs using distributed data parallelism on 8 GPUs with mixed precision. We use a per-GPU batch size of 16 and 8 data-loading workers. Data are drawn from the NAVSIM trainval split under the navtrain scene filter.

\subsection{Evaluation Results}

\begin{table*}[t]
\centering
\small
\setlength{\tabcolsep}{5pt}
\renewcommand{\arraystretch}{1.08}
\caption{\textbf{Comparison on NavTest.} 
Comparison of methods based on \emph{planning} paradigms and \emph{future reasoning}. \textbf{ProDrive} achieves the best overall performance, showing the efficacy of proactive planning and world model reranking. NC: no at-fault collision. DAC: drivable area compliance. TTC: time-to-collision. Comf.: comfort. EP: ego progress. PDMS: predictive driver model score.}
\label{tab:navsim_main}
\begin{tabular}{l|l|l|c c c c c >{\columncolor[gray]{0.92}}c}
\toprule
\textbf{Method} & \textbf{Planning} & \textbf{Future Reasoning Mechanism} & \textbf{NC}$\uparrow$ & \textbf{DAC}$\uparrow$ & \textbf{TTC}$\uparrow$ & \textbf{Comf.}$\uparrow$ & \textbf{EP}$\uparrow$ & \textbf{PDMS}$\uparrow$ \\
\midrule
Human             & --        & --                       & 100.0 & 100.0 & 100.0 & 99.9 & 87.5 & 94.8 \\
\midrule
AD-MLP  \cite{admlp}          & Reactive  & --                     & 93.0  & 77.3  & 83.6  & 100.0 & 62.8 & 65.6 \\
VADv2 \cite{VADv2}            & Reactive  & Rule-based Reranking     & 97.9  & 91.7  & 92.9  & 100.0 & 77.6 & 83.0 \\
UniAD \cite{UniAD}            & Reactive  & Rule-based Constraints   & 97.8  & 91.9  & 92.9  & 100.0 & 78.8 & 83.4 \\
LTF   \cite{LTF}            & Reactive  & --                     & 97.4  & 92.8  & 92.4  & 100.0 & 79.0 & 83.8 \\
PARA-Drive \cite{PARA_Drive}       & Reactive  & Rule-based Constraints   & 97.9  & 92.4  & 93.0  & 99.8  & 79.3 & 84.0 \\
TransFuser \cite{transfuser}       & Reactive  & --                     & 97.7  & 92.8  & 92.8  & 100.0 & 79.2 & 84.0 \\
LAW  \cite{LAW}             & Proactive & World-model Joint Training   & 96.4  & 95.4  & 88.7  & 99.9  & 81.7 & 84.6 \\
World4Drive  \cite{zheng2025world4driveendtoendautonomousdriving}             & Reactive & World-model Reranking   & 97.4  & 94.3  & 92.8  & 100.0  & 79.9 & 85.1 \\
FSDrive  \cite{zeng2025futuresightdrivethinkingvisuallyspatiotemporal}             & Proactive & Visual Spatio-Temporal CoT   & 98.2  & 93.8  & 93.3  & 99.9  & 80.1 & 85.1 \\
DRAMA  \cite{DRAMA}            & Reactive  & --                     & 98.0  & 93.1  & 94.8  & 100.0 & 80.1 & 85.5 \\
Epona  \cite{zhang2025eponaautoregressivediffusionworld}         & Reactive & World-model Joint Training   & 97.9  & 95.1  & 93.8  & 99.9  & 80.4 & 86.2 \\
Hydra-MDP  \cite{hydra_mdp}        & Reactive  & Model-free Reranking     & 98.3  & 96.0  & 94.6  & 100.0 & 78.7 & 86.5 \\
\midrule
ProDrive (Ours)   & Proactive & World-model Reranking    & \textbf{98.0}   & \textbf{95.4}   & \textbf{93.7}   & \textbf{99.9}   & \textbf{80.7}  & \textbf{86.6}  \\
\bottomrule
\end{tabular}
\end{table*}

\begin{figure*}[htbp]
    \centering
    
    \begin{subfigure}[b]{0.32\textwidth}
        \centering
        \includegraphics[width=\textwidth]{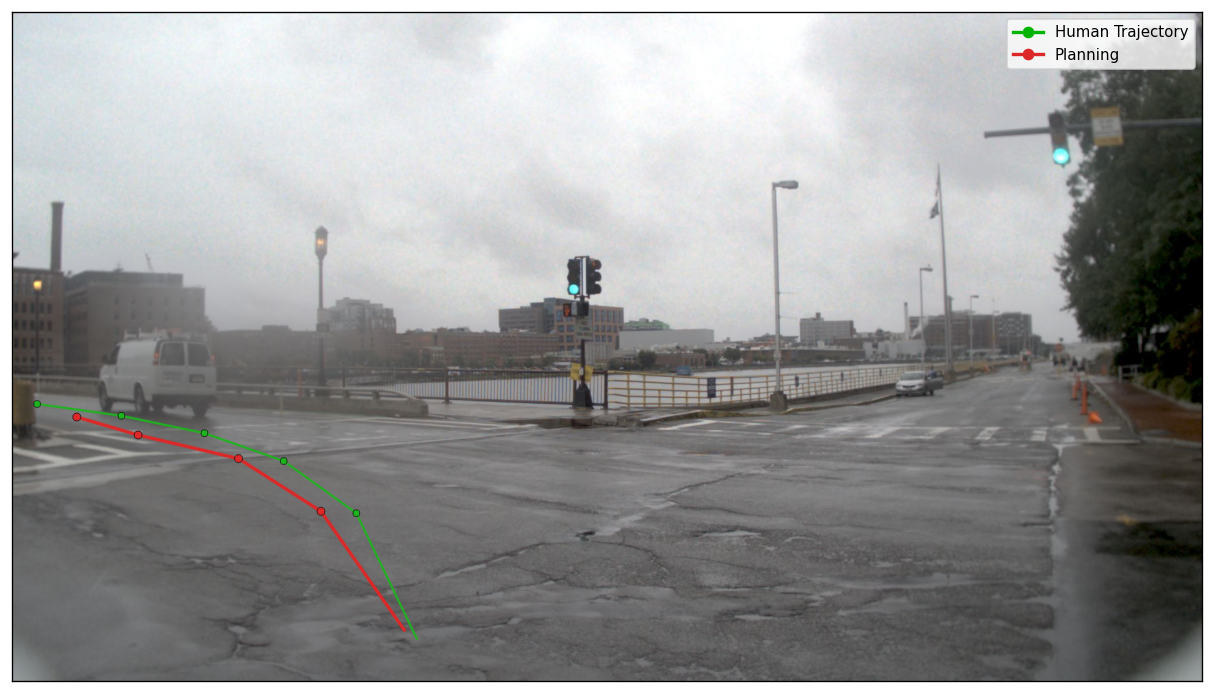} \\ 
        \vspace{4pt} 
        \includegraphics[width=\textwidth]{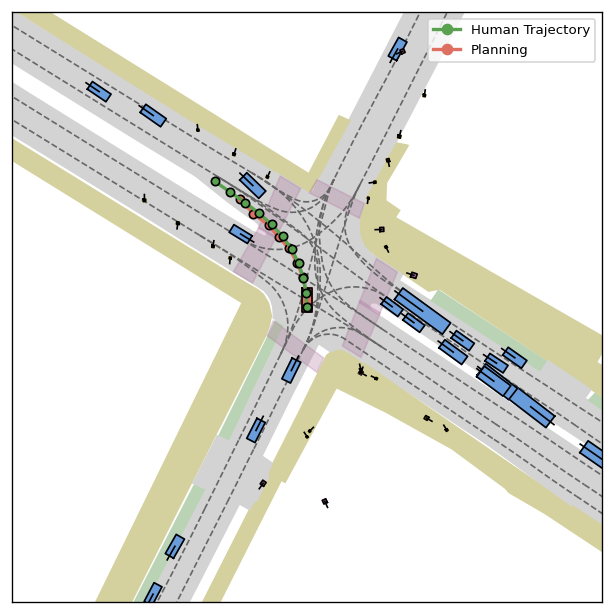}   
        \label{fig:sub_a}
    \end{subfigure}
    \hfill 
    \begin{subfigure}[b]{0.32\textwidth}
        \centering
        \includegraphics[width=\textwidth]{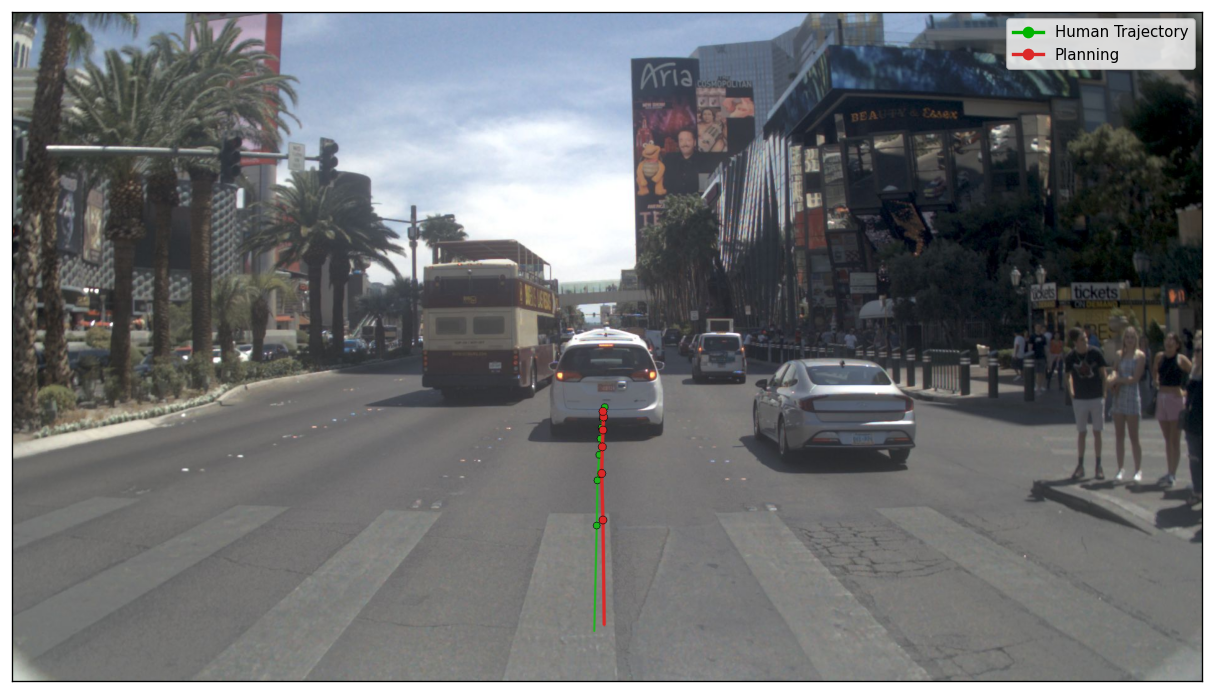} \\
        \vspace{4pt}
        \includegraphics[width=\textwidth]{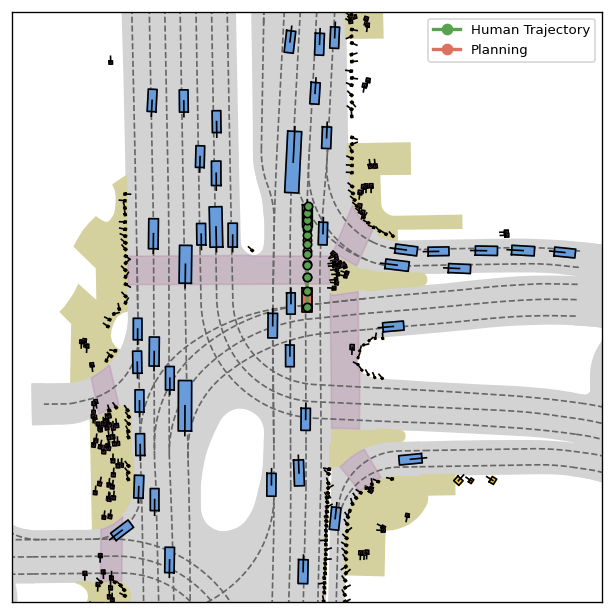}
        \label{fig:sub_b}
    \end{subfigure}
    \hfill
    \begin{subfigure}[b]{0.32\textwidth}
        \centering
        \includegraphics[width=\textwidth]{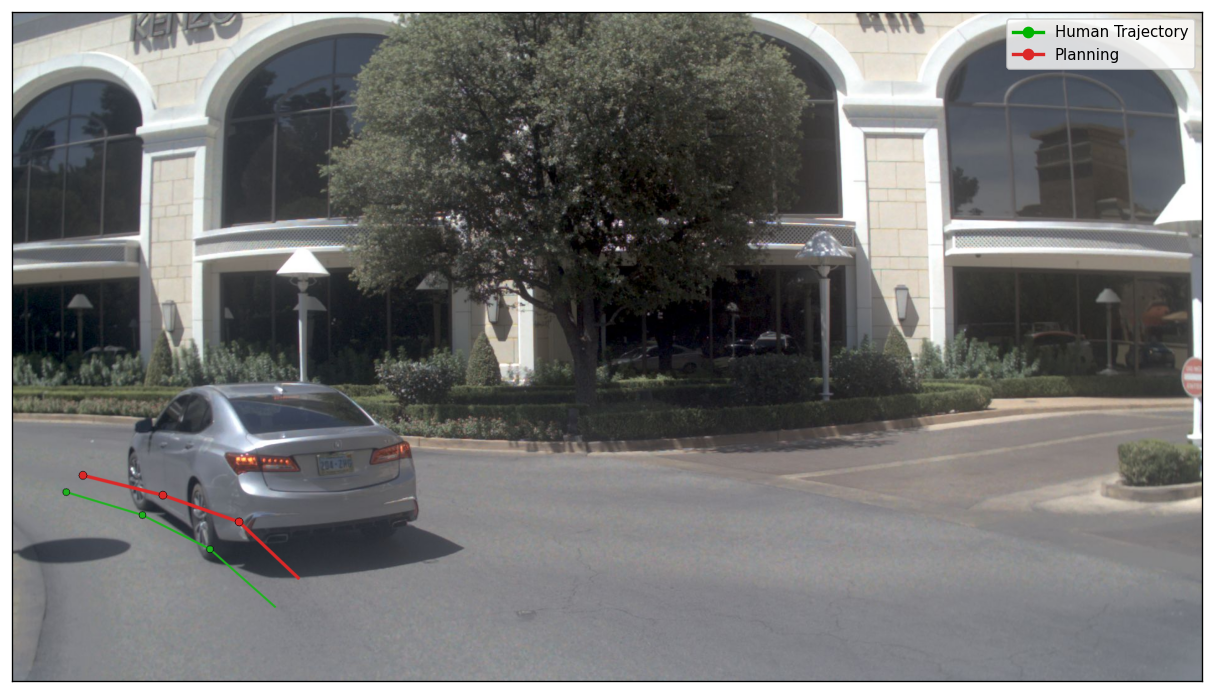} \\
        \vspace{4pt}
        \includegraphics[width=\textwidth]{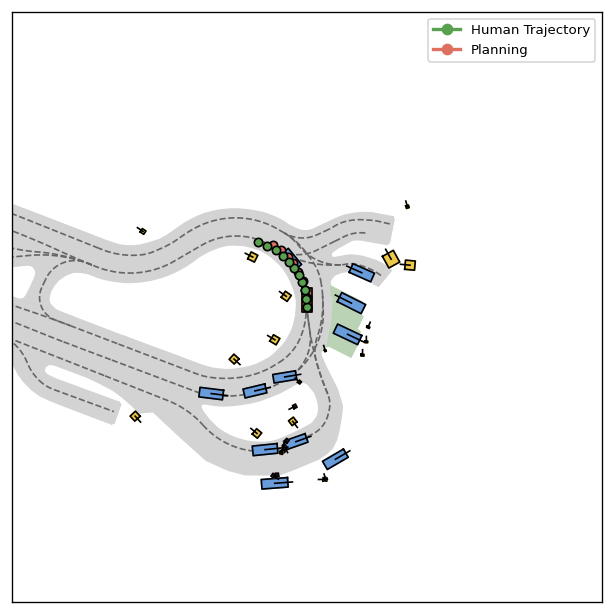}
        \label{fig:sub_c}
    \end{subfigure}

    \caption{\textbf{Qualitative examples of ProDrive.} 
    For each case, the top row shows the front-view observation with the planned trajectory overlaid, and the bottom row shows the corresponding bird's-eye-view scene with the predicted plan compared against the human trajectory. Across diverse driving scenarios, ProDrive produces safe and foresighted behaviors by anticipating the future motion of surrounding agents and capturing plausible interaction patterns. As a result, it can select effective trajectories while proactively avoiding potential collisions.}
    \label{fig:overall_label}
\end{figure*}

Table~\ref{tab:navsim_main} compares ProDrive with strong baselines on the NAVSIM test set. Benefiting from future-aware guidance provided by the world model, ProDrive enables the planner to acquire proactive planning capability and achieves strong safety performance, with \textbf{98.0} NC, \textbf{95.4} DAC, and \textbf{93.7} TTC. ProDrive also attains \textbf{80.7} EP, indicating improved long-horizon planning efficiency. Overall, our method outperforms all baselines across all reported metrics and achieves the best PDMS of \textbf{86.6}.

\subsection{Qualitative Analysis}

Qualitative results in Figure~\ref{fig:overall_label} further illustrate the advantage of ProDrive in complex interactive driving scenarios. Compared with reactive planners that make decisions primarily from the current observation, ProDrive is able to anticipate the future motion of surrounding vehicles and produce more foresighted planning behaviors, resembling the anticipatory reasoning of experienced human drivers. As a result, the planner tends to select trajectories that better balance safety and efficiency, proactively avoiding potential collision risks while maintaining desirable driving progress. These examples suggest that the planner is able to reason beyond the current scene and make effect decisions.

\subsection{Ablation Study}

To validate the necessity and effectiveness of the proposed design, we conduct three ablation studies:
\begin{itemize}
    \item w/o World Model: Remove the Environment Module and replace it with an MLP-based reward predictor for trajectory scoring and selection. 
    \item w/o Ego Token Injection: Remove the ego token injection mechanism, so the Ego and Environment modules are connected only through trajectory tokens.
    \item w/o Proactive Gradient: Block the reward loss from the Environment Module to the planner, removing the \emph{proactive} training signal from the Ego Module. 
\end{itemize}

The ablation results are summarized in Table~\ref{tab:ablation_main}. From these results, we draw the following three conclusions:

\paragraph{Effect of explicit future modeling.}
A possible counterargument to our design is that the Environment Module may act merely as a reward predictor, making explicit modeling of future scene evolution unnecessary. To test this hypothesis, we replace the world model with a $256\!\times\!1024\!\times\!1024\!\times\!6$ MLP that takes a trajectory token as input and directly predicts the reward for ranking candidate trajectories. As shown in Table~\ref{tab:ablation_main}, removing the world model leads to a substantial drop across all key metrics, with NC decreasing to 94.9, DAC to 93.9, and TTC to 90.8. These results indicate that explicit future-state modeling is critical for anticipating environmental hazards and improving safety. Moreover, the EP score drops to 79.5, suggesting that explicit future modeling also improves long-horizon planning efficiency. Overall, the PDMS falls to 83.5, significantly below the full model, which further confirms the importance of the Environment Module in the overall architecture.

\begin{table}[ht]
\centering
\small
\setlength{\tabcolsep}{4.5pt}
\renewcommand{\arraystretch}{1.08}
\caption{\textbf{Ablation study on NavTest.} We report the impact of removing the world model(WM),removing the ego token injection mechanism(ET) , and removing proactive feedback from the Environment Module to the Ego Module(PG). NC: no at-fault collision. DAC: drivable area compliance. TTC: time-to-collision. Comf.: comfort. EP: ego progress. PDMS: predictive driver model score.}
\label{tab:ablation_main}
\begin{tabular}{l | c c c c c >{\columncolor[gray]{0.92}}c}
\toprule
\textbf{Variant} & \textbf{NC}$\uparrow$ & \textbf{DAC}$\uparrow$ & \textbf{TTC}$\uparrow$ & \textbf{Comf.}$\uparrow$ & \textbf{EP}$\uparrow$ & \textbf{PDMS}$\uparrow$ \\
\midrule
w/o WM           & 94.9 & 93.9 & 90.8 & 99.9 & 79.5 & 83.5 \\
w/o ET   & 97.8 & 94.3 & 93.0 & 99.9 & 80.3 & 85.5 \\
w/o PG    & 97.6 & 94.9 & 93.1 & 99.9 & 80.3 & 85.8 \\
\midrule
ProDrive          & \textbf{98.0} & \textbf{95.4} & \textbf{93.7} & \textbf{99.9} & \textbf{80.7} & \textbf{86.6} \\
\bottomrule
\end{tabular}
\end{table}

\paragraph{Effect of planning-aware ego token injection.}
The refined ego tokens in ProDrive encode planning-relevant semantics accumulated through multiple rounds of query refinement. By injecting these tokens into the Environment Module, the world model is conditioned not only on the candidate trajectory itself, but also on richer planner-aware context for future prediction. The ablation results verify the importance of this design. As shown in Table~\ref{tab:ablation_main}, removing ego token injection reduces NC to 97.8, DAC to 94.3, TTC to 93.0, and EP to 80.3, indicating that without planner-aware semantic conditioning, the Environment Module becomes less effective at modeling future scene evolution and supporting safe, efficient planning. The overall PDMS also drops to 85.5, further confirming the contribution of ego token injection. In addition, we observe that the world-model-related losses in this ablation grow sharply after a period of training, suggesting that ego token injection is also important for stable collaborative optimization between the Ego and Environment modules.

\paragraph{Effect of proactive gradient feedback.}
In the final ablation, we preserve the full architecture and only block the gradient from the Environment Module to the planner, thereby removing the \emph{proactive} training signal while keeping all other components unchanged. This modification leads to a clear performance drop: as shown in Table~\ref{tab:ablation_main}, NC decreases to 97.6, DAC to 94.9, TTC to 93.1, and EP to 80.3, indicating that the planner's ability to make safe and effective decisions is noticeably weakened without future-aware gradient feedback. Together with the previous ablation on ego token injection, these results confirm the effectiveness of the proposed \textbf{Ego-Environment Coupling} mechanism. Removing either direction of interaction significantly degrades overall performance, whereas their joint presence enables the Ego and Environment modules to collaborate effectively and achieve \emph{ego-environment co-evolution}.

\section{Conclusion}
\label{sec:conclusion}

We presented \textbf{ProDrive}, a world-model-based proactive planning framework for end-to-end autonomous driving. ProDrive moves beyond current-observation-driven reactive planning by tightly coupling trajectory generation and future scene prediction through \emph{ego-environment co-evolution}. By injecting planner-refined ego tokens into the world model and propagating future-aware rewards back to the planner through end-to-end optimization, ProDrive enables planning to be shaped by anticipated scene evolution. Experiments on NAVSIM show consistent gains in safety and planning efficiency over strong baselines, while ablations confirm the importance of explicit future modeling, ego token injection, and proactive gradient feedback, especially in challenging interactive driving scenarios. Future work includes extending ProDrive to more expressive world models. We hope our work serves as a useful step toward autonomous driving systems in which planning is not merely reactive to the current observation, but is proactively guided by learned predictions of future scene evolution.


\newpage
{
    \small
    \bibliographystyle{ieeenat_fullname}
    \bibliography{main}
}


\end{document}